\documentclass[10pt,twocolumn,letterpaper]{article}

\usepackage{iccv}
\usepackage{times}
\usepackage{graphicx}
\usepackage{epstopdf}
\usepackage{amsmath}
\usepackage{amssymb}

\usepackage{subfigure}
\usepackage{multirow}
\usepackage{url}

\usepackage[pagebackref=true,breaklinks=true,letterpaper=true,colorlinks,bookmarks=false]{hyperref}

\iccvfinalcopy 

\def\iccvPaperID{2185} 

\ificcvfinal\fi

\begin{document}
	
\title{Deep Convolutional Neural Networks with Merge-and-Run Mappings}

\author{
Liming Zhao$^1$
\quad
Jingdong Wang$^2$
\quad
Xi Li$^1$
\quad
Zhuowen Tu$^3$
\quad
Wenjun Zeng$^2$
\\
$^1$Zhejiang University
\quad
$^2$Microsoft Research
\quad
$^3$UC San Diego\\
{\tt\small \{zhaoliming,xilizju\}@zju.edu.cn}
\quad
{\tt\small \{jingdw,wezeng\}@microsoft.com}
\quad
{\tt\small ztu@ucsd.edu}
}

\maketitle
\thispagestyle{empty}


\begin{abstract}
A deep
residual network,
built by stacking a sequence of residual blocks,
is easy to train,
because identity mappings skip residual branches
and thus improve information flow.
To further reduce the training difficulty,
we present a simple network architecture,
\emph{deep merge-and-run neural networks}.
The novelty lies in
a modularized building block,
\emph{merge-and-run block},
which assembles residual branches in parallel
through a \emph{merge-and-run mapping}:
Average the inputs of these residual branches (\emph{Merge}),
and add the average
to the output of each residual branch
as the input of the subsequent residual branch (\emph{Run}),
respectively.
We show that the merge-and-run mapping is
a linear idempotent function
in which
the transformation matrix is idempotent,
and thus improves information flow, making training easy.
In comparison to residual
networks,
our networks enjoy compelling advantages:
they contain much shorter paths,
and the width, i.e., the number of channels, is increased.
We evaluate the performance on the standard recognition tasks.
Our approach demonstrates consistent improvements
over ResNets with the comparable setup,
and achieves competitive results
(e.g., $3.57\%$ testing error on CIFAR-$10$,
$19.00\%$ on CIFAR-$100$,
$1.51\%$ on SVHN).
\end{abstract}

\section{Introduction}

Deep convolutional neural networks,
since the breakthrough result
in the ImageNet classification challenge~\cite{AlexNet},
have been widely studied~\cite{GoogLeNet, SimonyanZ14a, ResNet}.
Surprising performances have been achieved in many other computer vision tasks,
including object detection~\cite{GirshickDDM14},
semantic segmentation~\cite{LongSD15},
edge detection~\cite{XieT15}, and so on.

Residual networks (ResNets)~\cite{ResNet} have been attracting a lot of attentions
since it won the ImageNet challenge
and various extensions
have been studied~\cite{WideResNet, TargAL16, ResOfRes, MultiRes}.
The basic unit
is a residual block consisting
of a residual branch
and an identity mapping.
Identify mappings
introduce short paths
from the input to the intermediate layers
and from the intermediate layers to the output layers~\cite{WangWZZ16, VeitWB16},
and thus reduce the training difficulty.

\begin{figure}[t]
\centering
\footnotesize
{(a)~\includegraphics[scale=0.33]{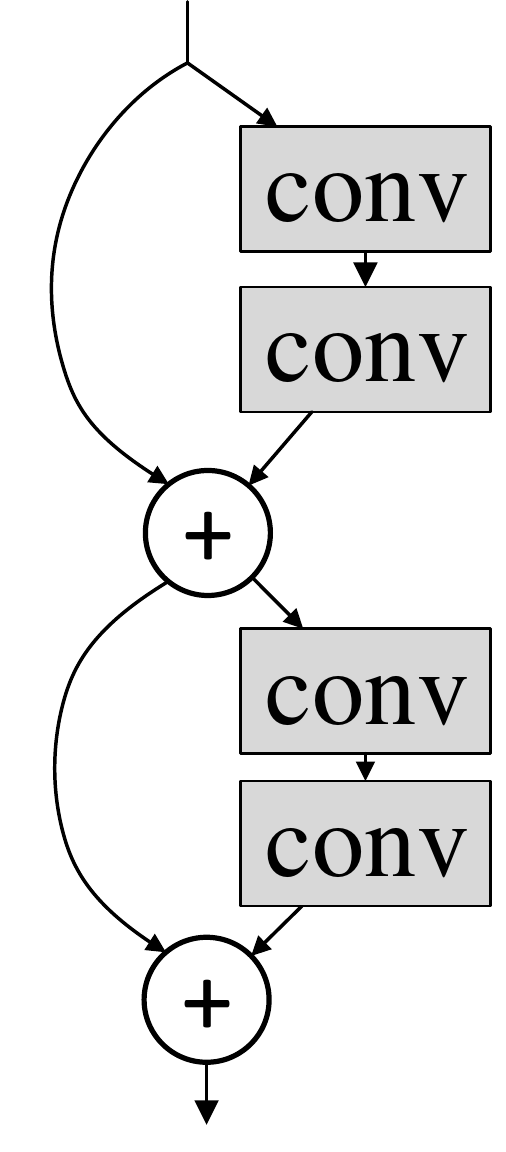}}~~~~~~
{(b)~\includegraphics[scale=0.33]{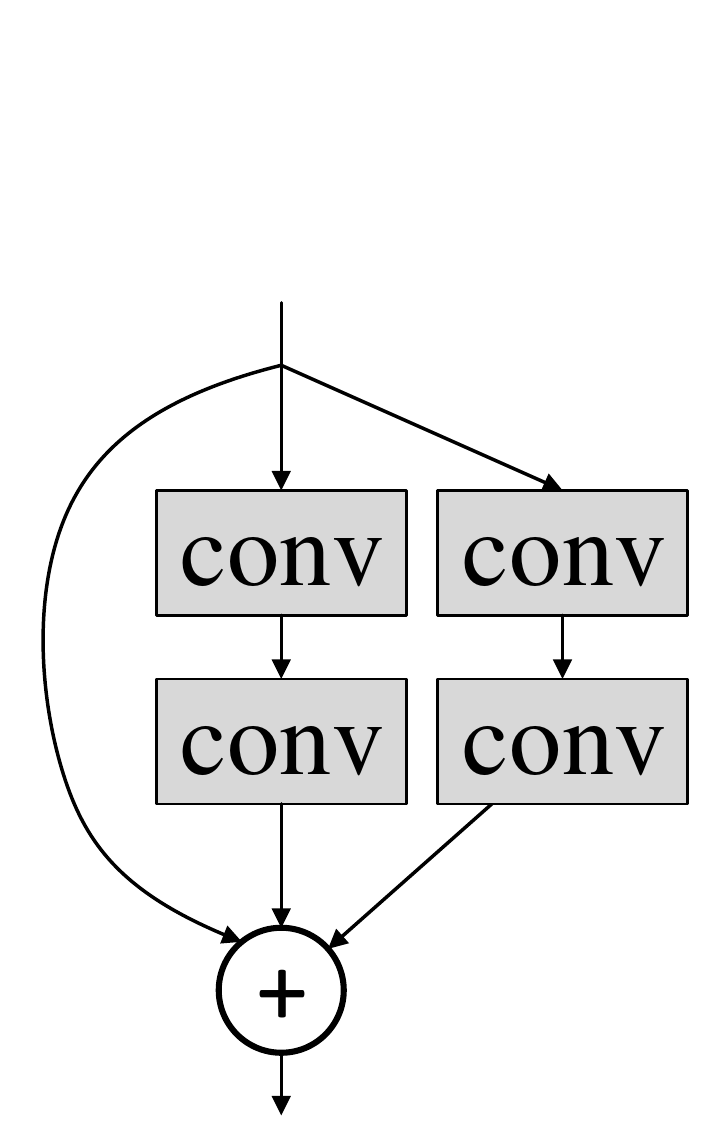}}~~~~~~
{(c)~\includegraphics[scale=0.33]{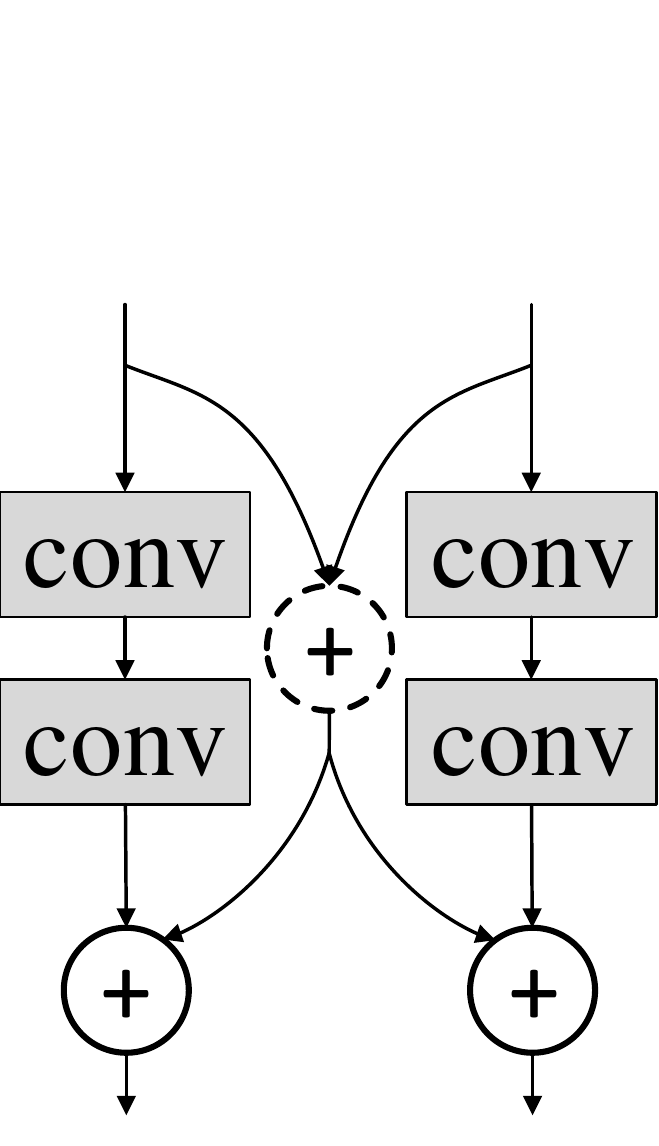}}

\caption{Illustrating the building blocks:
(a) Two residual blocks;
(b) An inception-like block;
(c) A merge-and-run block.
(a) corresponds to two blocks in ResNets
and assembles two residual branches sequentially
while (b) and (c) both assemble the same two residual branches in parallel.
(b) and (c) adopt two different skip connections:
identity mappings and our proposed merge-and-run mappings.
The dot circle denotes the average operation,
and the solid circle denotes the sum operation.}
\label{fig:networkblocks}
\vspace{-.3cm}
\end{figure}

In this paper,
we are interested in further reducing the training difficulty
and present a simple network architecture,
called deep merge and run neural networks,
which assemble residual branches more effectively.
The key point is a novel building block,
the \emph{merge-and-run} block,
which assembles residual branches
in parallel
with a \emph{merge-and-run mapping}:
Average the inputs of these residual branches (\emph{Merge}),
and add the average
to the output of each residual branch
as the input of the subsequent residual branch
(\emph{Run}), respectively.
Figure~\ref{fig:networkblocks}
depicts the architectures
by taking two residual branches as an example:
(a) two residual blocks
that assemble two residual branches sequentially,
(b) an inception-like block
and (c) a merge and run block
both assemble the same two residual branches in parallel.

Obviously, the resulting network contains shorter paths
as the parallel assembly of residual branches
directly reduces the network depth.
We give a straightforward verification:
The average length of two residual blocks is $2$,
while the average lengths of the corresponding inception-like block and merge-and-run block
are $\frac{4}{3}$ and $\frac{2}{3}$, respectively.
Our networks, built by stacking merge-and-run blocks,
are less deep and thus easier to train.

We show that
the merge-and-run mapping
is a linear idempotent function,
where the transformation matrix is idempotent.
This implies that
the information from the early blocks
can quickly flow to the later blocks,
and the gradient can be quickly back-propagated
to the early blocks from the later blocks.
This point essentially provides a theoretic counterpart
of short paths, showing
the training difficulty is reduced.

We further show that
merge-and-run blocks are
wider than residual blocks.
Empirical results validate that
for very deep networks,
as a way to increase the number of layers,
increasing the width
is more effective
than increasing the depth.
Besides,
we discuss the generalizability of merge-and-run mappings
to other linear idempotent transformations,
and the extension to more residual branches.

The empirical results demonstrate
that the performances of our networks are superior
to the corresponding ResNets
with comparable setup
on CIFAR-$10$, CIFAR-$100$, SVHN and ImageNet.
Our networks achieve
competitive results
compared to state-of-the-arts
(e.g., $3.57\%$ testing error on CIFAR-$10$,
$19.00\%$ on CIFAR-$100$,
$1.51\%$ on SVHN).

\section{Related Works}

There have been rapid and great progress
of deep neural networks
in various aspects,
such as optimization techniques~\cite{SutskeverMDH13,IoffeS15,LoshchilovH16a}, initialization schemes~\cite{MishkinM15},
regularization strategies~\cite{SrivastavaHKSS14},
activation and pooling functions~\cite{MSRAinit,ClevertUH15},
network architecture~\cite{LinCY13,VGGnet,PezeshkiFBCB16,smith2016deep},
and applications.
In particular,
recently network architecture design
has been attracting a lot of attention.


Highway networks~\cite{SimonyanZ14a},
residual networks~\cite{ResNet,ResNetV2}, and GoogLeNet~\cite{GoogLeNet}
are shown to
be able to effectively train a very deep
(over $40$ and even hundreds or thousands) network.
The identity mapping or the bypass path are thought
as the key factor to
make the training of very deep networks easy.
Following the ResNet architecture,
several variants are developed
by modifying the architectures,
such as wide residual networks~\cite{WideResNet},
ResNet in ResNet~\cite{TargAL16},
multilevel residual networks~\cite{ResOfRes},
multi-residual networks~\cite{MultiRes}, and so on.
Another variant, DenseNets~\cite{HuangLW16a},
links all layers with identity mappings
and are able to improve the effectiveness
through feature reuse.
In addition,
optimization techniques,
such as stochastic depth~\cite{StochasticDepth}
for ResNet optimization,
are developed.

Deeply-fused networks~\cite{WangWZZ16},
FractalNet~\cite{LarssonMS16a}, and ensemble view~\cite{VeitWB16}
point out
that a ResNet and a GoogleNet~\cite{GoogLeNet,InceptionV3,InceptionV4}
are a mixture
of many dependent networks.
Ensemble view~\cite{VeitWB16}
observes that ResNets
behave like an exponential ensemble
of relatively shallow networks,
and points out that introducing short paths
helps ResNets to avoid the vanishing gradient problem,
which is similar to the analysis
in deeply-fused networks~\cite{WangWZZ16} and
FractalNet~\cite{LarssonMS16a}.

The architecture of our approach is closely related to
Inception~\cite{GoogLeNet}
and Inception-ResNet blocks~\cite{InceptionV4},
multi-residual networks~\cite{MultiRes},
and ResNeXt~\cite{XieGDTH16},
which also contain multiple branches in each block.
One notable point is that
we introduce merge-and-run mappings,
which are linear idempotent functions,
to improve information flow
for building blocks consisting of parallel residual branches.

In comparison to ResNeXts~\cite{XieGDTH16} that also assemble
residual branches in parallel,
our approach adopts parallel assembly to directly reduce the depth
and does not modify residual branches,
while ResNeXts~\cite{XieGDTH16}
transform a residual branch to many small residual branches.
Compared to Inception~\cite{GoogLeNet} and Inception-ResNet blocks~\cite{InceptionV4}
that are highly customized,
our approach requires less efforts to design
and more flexible.


\begin{figure}[t]
\centering
\footnotesize
{(a)~\includegraphics[height=1.3\linewidth, angle=0]{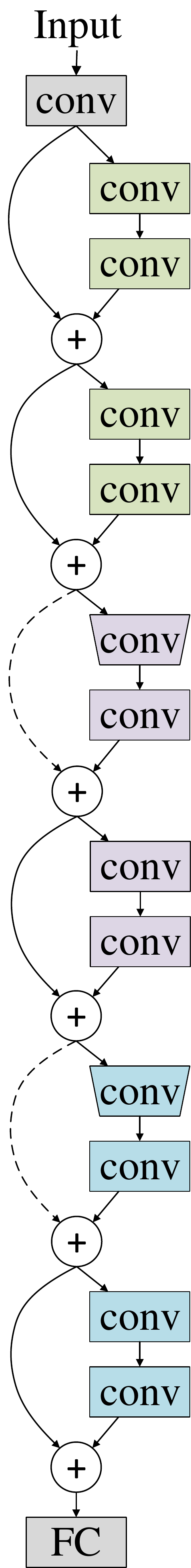}}~~~~~~~~~~~~~
{(b)~\includegraphics[height=1.3\linewidth, angle=0]{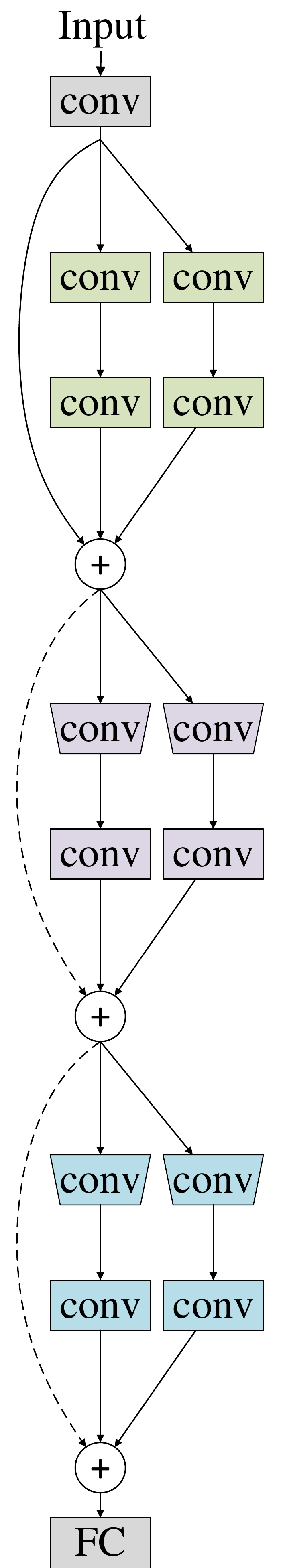}}~~~~~~~~~~
{(c)~\includegraphics[height=1.3\linewidth, angle=0]{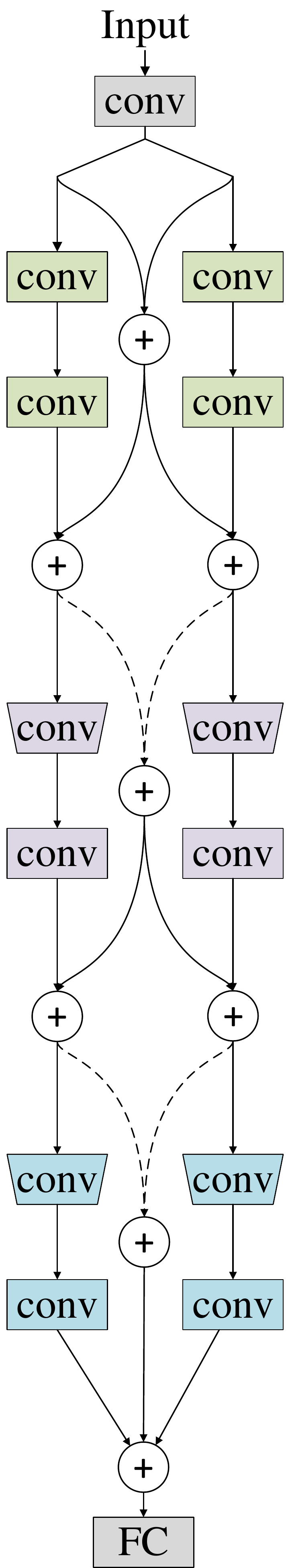}}
\caption{(a) a deep residual network;
(b) a network built by stacking inception-like blocks;
(c) our deep merge-and-run neural network
built by stacking merge-and-run blocks.
The trapezoid shape indicates that down-sampling
occurs in the corresponding layer, and
the dashed line denotes
a projection shortcut as in~\cite{ResNet}. }
\label{fig:networkexamples}\vspace{-.4cm}
\end{figure}
\section{Deep Merge-and-Run Neural Networks}

\newcommand{\blocka}[2]{\multirow{3}{*}{\(\left[\begin{array}{c}\text{3$\times$3, #1}\\[-.1em] \text{3$\times$3, #1} \end{array}\right]\)$\times$#2}
}
\newcommand{\blockb}[3]{\multirow{3}{*}{\(\left[\begin{array}{c}\text{1$\times$1, #2}\\[-.1em] \text{3$\times$3, #2}\\[-.1em] \text{1$\times$1, #1}\end{array}\right]\)$\times$#3}
}
\renewcommand\arraystretch{1.1}
\setlength{\tabcolsep}{3pt}

\begin{table*}[t]
\setlength{\tabcolsep}{12pt}
	\caption{Network architectures.
 Inside the brackets are the shape of the residual,
 inception-like and merge-and-run blocks,
 and outside the brackets is the number of stacked blocks on a stage.
Downsampling is performed in conv2\_1, and conv3\_1 with stride 2.
As done~\cite{ResNet},
we use $1\times 1$ convolutions to replace identity mappings for ResNets and Inception-like blocks,
and perform $1\times 1$ convolutions before merging in merge-and-run blocks
when the widths increase across stages.
In each convolution, the input channel number can be known from the preceding layer.
}
\label{tab:networtdescriptions}
	\centering
			\footnotesize
			\begin{tabular}{c|c|c|c}
				\hline
				Layers & Output size & ResNets &  DMRNets/DILNets   \\
				\hline
				conv$0$ & $32 \times 32$  & \multicolumn{2}{c}{$ \operatorname{conv}(3 \times3, 16)$}  \\
				\hline
				con1$\_x$ &  $ 32 \times 32$
                                  & $\begin{bmatrix} \operatorname{conv}(3\times 3, 16) \\
                                  \operatorname{conv}(3\times 3, 16)  \end{bmatrix}
                                  \times \frac{2L}{3}$
                                  &
				$\begin{bmatrix} \operatorname{conv}(3\times 3, 16) & \operatorname{conv}(3\times 3, 16) \\
                                  \operatorname{conv}(3\times 3, 16) & \operatorname{conv}(3\times 3, 16) \end{bmatrix}
                                  \times \frac{L}{3}$
                                  \\

				\hline
				conv$2\_x$ &  $16 \times 16 $
                                   &  $\begin{bmatrix} \operatorname{conv}(3\times 3, 32) \\
                                  \operatorname{conv}(3\times 3, 32)  \end{bmatrix}
                                  \times \frac{2L}{3}$
                                  &
				$\begin{bmatrix} \operatorname{conv}(3\times 3, 32) & \operatorname{conv}(3\times 3, 32) \\
                                  \operatorname{conv}(3\times 3, 32) & \operatorname{conv}(3\times 3, 32) \end{bmatrix}
                                  \times \frac{L}{3}$  \\

				\hline
				conv$3\_x$ &  $8 \times 8$
                                  &
                                  $\begin{bmatrix} \operatorname{conv}(3\times 3, 64) \\
                                  \operatorname{conv}(3\times 3, 64)  \end{bmatrix}
                                  \times \frac{2L}{3}$
                                  &
				$\begin{bmatrix} \operatorname{conv}(3\times 3, 64) & \operatorname{conv}(3\times 3, 64) \\
                                  \operatorname{conv}(3\times 3, 64) & \operatorname{conv}(3\times 3, 64) \end{bmatrix}
                                  \times \frac{L}{3}$  \\
				\hline
				Classifier & 1$\times$1  & \multicolumn{2}{c}{average pool, FC, softmax} \\
				\hline
			\end{tabular}
\end{table*}

\subsection{Architectures}
We introduce the architectures
by considering a simple realisation,
assembling two residual branches
in parallel
to form the building blocks.
We first introduce the building blocks in ResNets,
then a straightforward manner
to assemble residual branches in parallel,
and finally our building blocks.

The three building blocks
are illustrated in Figure~\ref{fig:networkblocks}.
Examples of the corresponding network structures,
ResNets,
DILNets (\underline{d}eep \underline{i}nception-\underline{l}ike neural networks),
and DMRNets (\underline{d}eep \underline{m}erge-and-\underline{r}un neural networks),
are illustrated in
Figure~\ref{fig:networkexamples}.
The descriptions of network structures used in this paper
are given in Table~\ref{tab:networtdescriptions}.

\vspace{.1cm}
\noindent\textbf{Residual blocks.}
A residual network is composed of
a sequence of residual blocks.
Each residual block contains two branches:
identity mapping and residual branch.
The corresponding function is given as,
\begin{align}
\mathbf{x}_{t+1}
= H_t(\mathbf{x}_t) + \mathbf{x}_t.
\label{eqn:residualblock}
\end{align}
Here, $\mathbf{x}_t$ denotes the input of the $t$th residual block.
$H_t(\cdot)$ is a transition function,
corresponding to the residual branch
composed of a few stacked layers.

\vspace{.5em}
\noindent\textbf{Inception-like blocks.}
We assemble two residual branches in parallel
and sum up the outputs from the two residual branches
and the identity mapping.
The functions,
corresponding to the $(2t)$th and $(2t+1)$th residual branches,
are as follows,
\begin{align}
\mathbf{x}_{2(t+1)} = H_{2t}(\mathbf{x}_{2t}) + H_{2t+1}(\mathbf{x}_{2t}) + \mathbf{x}_{2t},
\label{eqn:inceptionlikeblock}
\end{align}
where $\mathbf{x}_{2t}$ and $\mathbf{x}_{2(t+1)}$
are the input and the output of the $t$th inception-like block.
This structure resembles the building block in
the concurrently-developed ResNeXt~\cite{XieGDTH16},
but the purposes are different:
Our purpose is to reduce the depth through assembling residual branches
in parallel
while the purpose of ResNeXt~\cite{XieGDTH16}
is to transform a single residual branches to many small residual branches.

\vspace{.5em}
\noindent\textbf{Merge-and-run.}
A {merge-and-run} block is
formed by assembling two residual branches
in parallel
with a \emph{merge-and-run} mapping:
Average the inputs of two residual branches (\emph{Merge}),
and add the average
to the output of each residual branch
as the input of the subsequent residual branch
(\emph{Run}), respectively.
It is formulated as below,
\begin{align}
\mathbf{x}_{2(t+1)} =~& H_{2t}(\mathbf{x}_{2t}) + \frac{1}{2} (\mathbf{x}_{2t}
+ \mathbf{x}_{2t + 1}), \nonumber\\
\mathbf{x}_{2(t+1)+1} =~& H_{2t+1}(\mathbf{x}_{2t+1}) + \frac{1}{2} (\mathbf{x}_{2t}
+ \mathbf{x}_{2t + 1}),
\label{eqn:mergeandrunblock}
\end{align}
where $\mathbf{x}_{2t}$
and $\mathbf{x}_{2t+1}$
($\mathbf{x}_{2(t+1)}$ and $\mathbf{x}_{2(t+1)+1}$)
are the inputs (outputs) of two residual branches
of the $t$th block.
There is a clear difference
from inception-like blocks in Equation~\ref{eqn:inceptionlikeblock}:
the inputs of two residual branches are different,
and their outputs are also separated.

\begin{figure}[t]
	\centering
	\includegraphics[width=.49\linewidth]{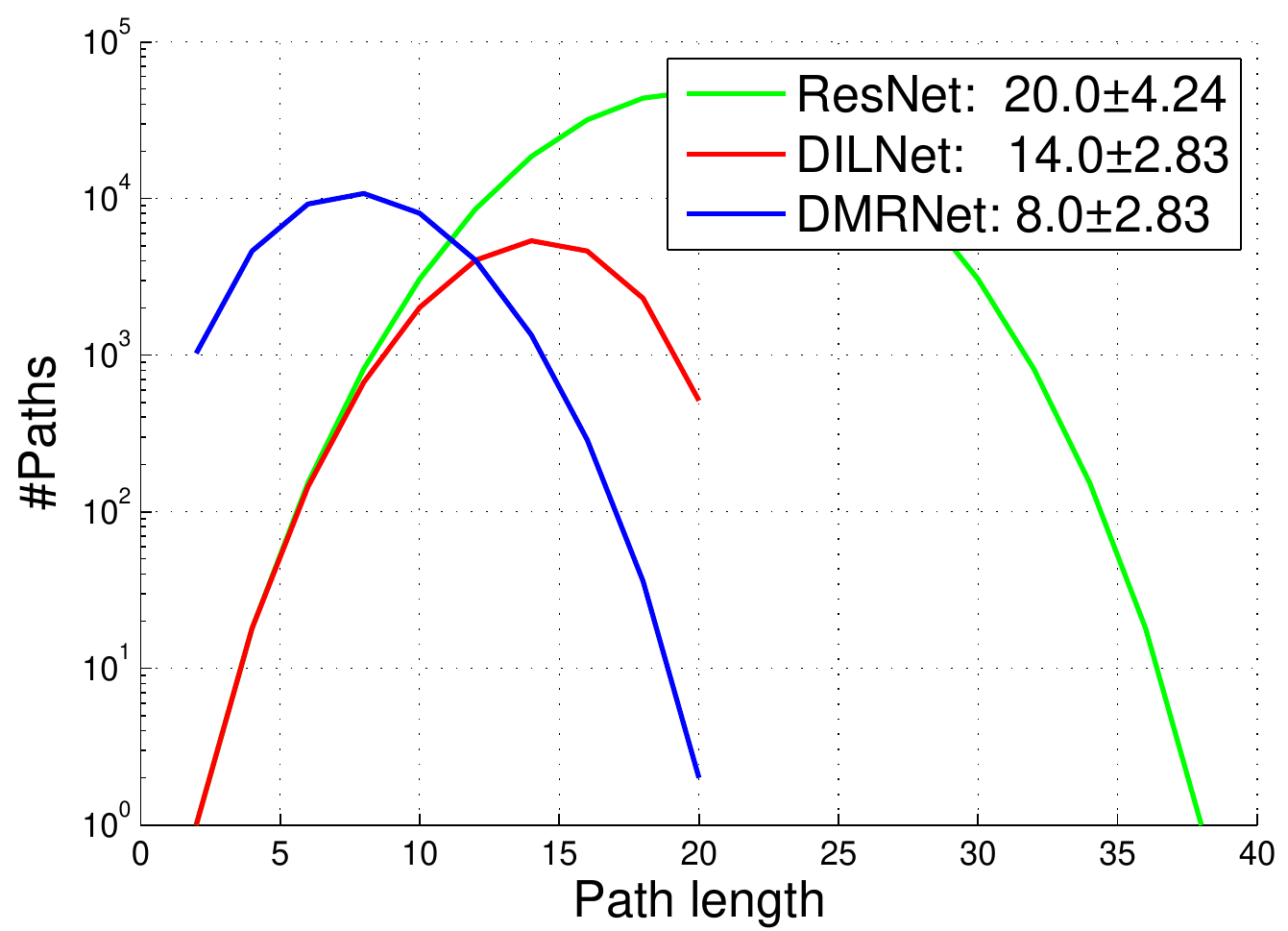}~~~
\includegraphics[width=.49\linewidth]{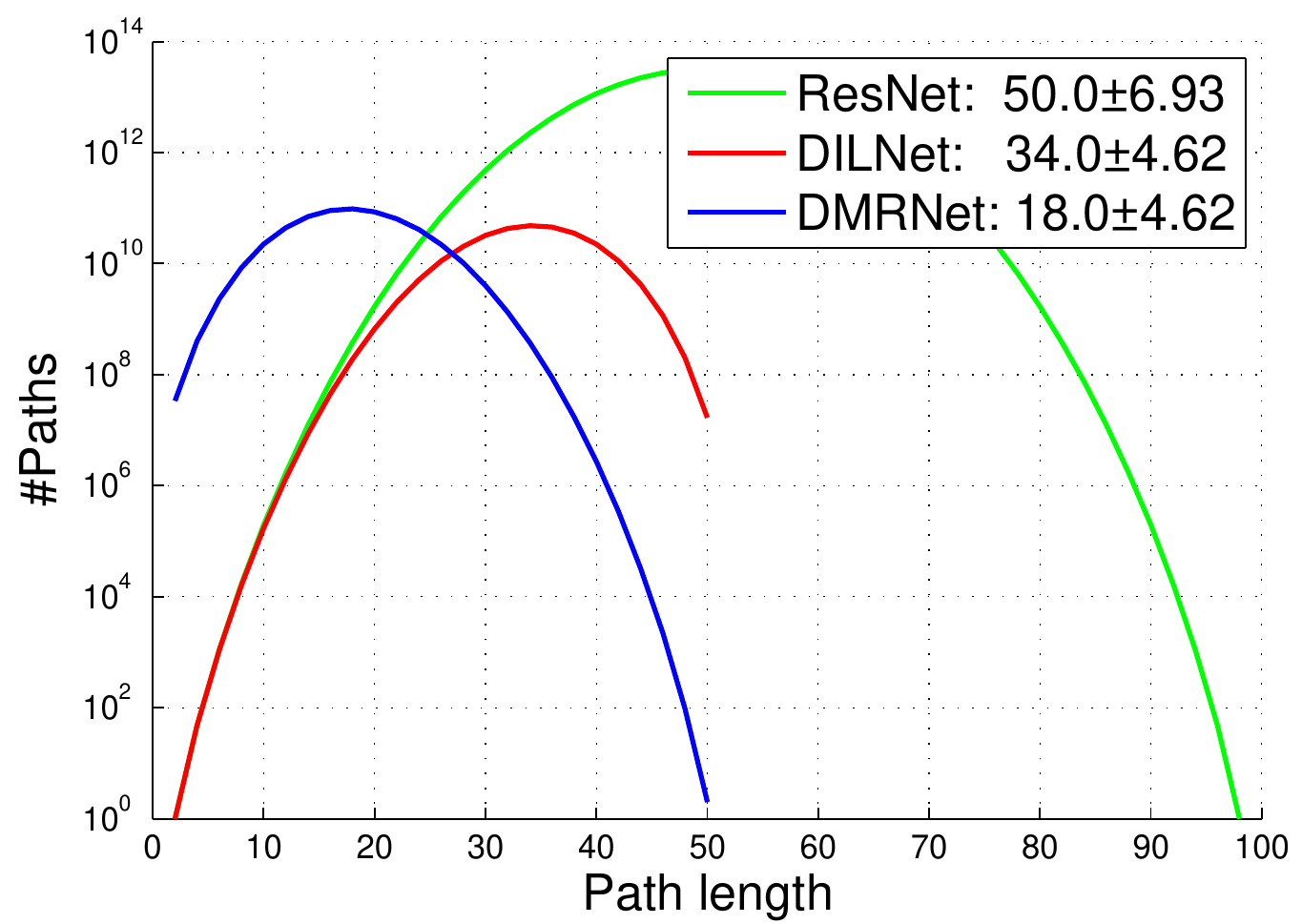}
	\caption{Comparing the distributions
		of the path lengths for three networks.
		Different networks: (avg length $\pm$ std).
        Left: $L=9$. Right: $L=24$.}
	\label{fig:depthdistributions}\vspace{-.4cm}
\end{figure}

\subsection{Analysis}
\label{sec:analysis}

\noindent\textbf{Information flow improvement.}
We transform Equation~\ref{eqn:mergeandrunblock}
into the matrix form,
\begin{align}
\begin{bmatrix}
\mathbf{x}_{2(t+1)} \\[0.3em]
\mathbf{x}_{2(t+1)+1}
\end{bmatrix}
=
\begin{bmatrix}
 H_{2t}(\mathbf{x}_{2t}) \\[0.3em]
 H_{2t+1}(\mathbf{x}_{2t+1})
 \end{bmatrix}
 + \frac{1}{2}\begin{bmatrix}
 \mathbf{I} & \mathbf{I} \\[0.3em]
 \mathbf{I} & \mathbf{I}
   \end{bmatrix}
   \begin{bmatrix}
   \mathbf{x}_{2t} \\[0.3em]
   \mathbf{x}_{2t+1}
   \end{bmatrix},
    \label{eqn:mergeandrunblockforwidth}
\end{align}
where $\mathbf{I}$ is an $d\times d$ identity matrix
and $d$ is the dimension of $\mathbf{x}_{2t}$ (and $\mathbf{x}_{2t+1}$).
$\mathbf{M} =
\frac{1}{2}\begin{bmatrix}
 \mathbf{I} & \mathbf{I} \\[0.3em]
 \mathbf{I} & \mathbf{I}
   \end{bmatrix}$
is the transformation matrix
of the merge-and-run mapping.

It is easy to show that like the identity matrix $\mathbf{I}$,
$\mathbf{M}$
is an~\emph{idempotent} matrix,
i.e., $\mathbf{M}^n = \mathbf{M}$,
where $n$ is an arbitrary positive integer.
Thus, we have\footnote{Similar to identity mappings,
the analysis is applicable to
the case that the flow
is not stopped
by nonlinear activation ReLU.
This equation is similar
to the derivation with identity mappings
in~\cite{ResNetV2}.}

\begin{align}
\begin{bmatrix}
\mathbf{x}_{2(t+1)} \\[0.3em]
\mathbf{x}_{2(t+1)+1}
\end{bmatrix}
=~&
\begin{bmatrix}
 H_{2t}(\mathbf{x}_{2t}) \\[0.3em]
 H_{2t+1}(\mathbf{x}_{2t+1})
 \end{bmatrix} + \nonumber\\ 
 &
\mathbf{M}
\sum_{i=t'}^{t-1}
\begin{bmatrix}
 H_{2i}(\mathbf{x}_{2i}) \\[0.3em] 
 H_{2i+1}(\mathbf{x}_{2i+1})
 \end{bmatrix}
+ \mathbf{M}\begin{bmatrix}
\mathbf{x}_{2t'} \\[0.3em]
\mathbf{x}_{2t' + 1}
\end{bmatrix},
\label{eqn:informationflowimprovement}
\end{align}
where $t' < t$ corresponds to
an earlier block\footnote{The second term in the right-hand side of Equation~\ref{eqn:informationflowimprovement}
does not exist if $(t+1)$ corresponds to the block right after the input of the whole network.}.
This implies that during the forward propagation
there are quick paths directly
sending the input and the outputs of
the intermediate residual branches
to the later block.
We have a similar conclusion
for gradient back-propagation.
Consequently,
merge-and-run mappings
improve both forward and backward information flow.

\begin{figure}[t]
\centering
\footnotesize
{(a)~\includegraphics[scale=.45]{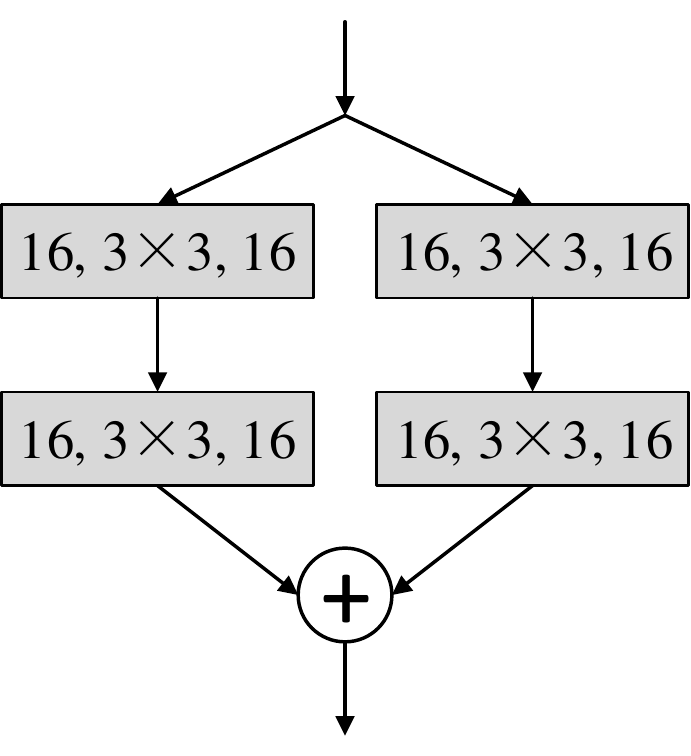}}~~~~~~~
{(b)~\includegraphics[scale=.45]{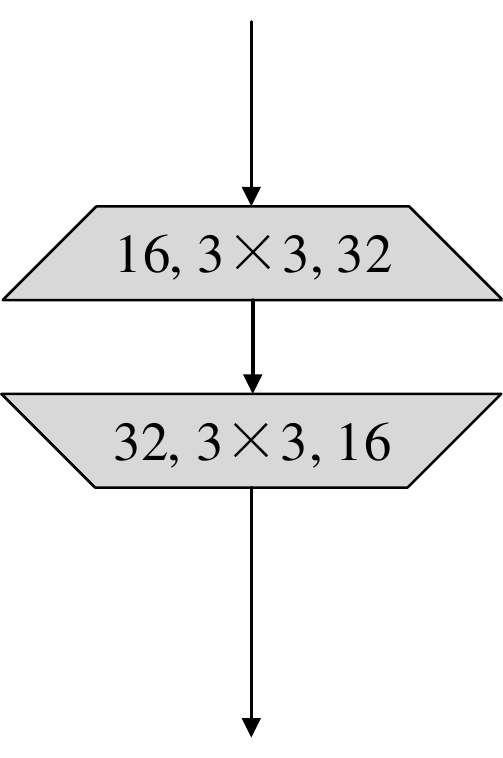}}
\caption{Illustrating the two residual branches shown in (a)
are transformed to a single residual branch shown in (b).
(a) All $4$ convolutions are $(16, 3 \times 3, 16)$.
(b) The $2$ convolutions are $(16, 3 \times 3, 32)$ and $(32, 3 \times 3, 16)$,
from narrow ($16$) to wide ($32$), and then from wide ($32$)
back to narrow ($16$).}
\label{fig:wideinceptionlike}\vspace{-.4cm}
\end{figure}

\vspace{.5em}
\noindent\textbf{Shorter paths.}
All the three networks
are mixtures of paths,
where a path is defined
as a sequence of connected residual branches,
identity mappings,
and possibly other layers (e.g., the first convolution layer,
the FC layer)
from the input to the output.
Suppose each residual branch contains $B$ layers
(there are $2$ layers for the example shown in Figure~\ref{fig:networkblocks}),
and the ResNet,
DIRNet
and DMRNet contain $2L$,
$L$, and $L$ building blocks,
the average lengths (without counting projections in short-cut connections)
are
$BL+2$,
$\frac{2B}{3}L+2$,
and $\frac{B}{3}L+2$, respectively.
Figure~\ref{fig:depthdistributions}
shows the distributions
of path lengths
of the three networks.
Refer to Table~\ref{tab:networtdescriptions}
for the details of the network structures.

It is shown in~\cite{ResNet, SimonyanZ14a}
that for very deep networks
the training becomes hard
and that a shorter (but still very deep) plain network
performs even better than a longer plain network\footnote{Our empirical results even
show that the deepest path in ResNets hurts the training of other paths
and thus deteriorates the performance.}.
According to Figure~\ref{fig:depthdistributions}
showing that
the lengths of the paths in
our proposed network are distributed
in the range of lower lengths,
the proposed deep merge-and-run network
potentially performs better.

\vspace{.5em}
\noindent\textbf{Inception-like blocks are wider.}
We rewrite Equation~\ref{eqn:inceptionlikeblock}
in a matrix form,
\begin{align}
\mathbf{x}_{2(t+1)} =
\begin{bmatrix}
\mathbf{I} & \mathbf{I}
\end{bmatrix}
\begin{bmatrix}
 H_{2t}(\mathbf{x}_{2t}) \\[0.3em]
 H_{2t+1}(\mathbf{x}_{2t})
 \end{bmatrix}
  + \mathbf{x}_{2t}.
  \label{eqn:inceptionlikeblockforwidth}
\end{align}
Considering the two parallel residual branches,
i.e., the first term of the right-hand side,
we have several observations.
(1) The intermediate representation,
$\begin{bmatrix}
 H_{2t}(\mathbf{x}_{2t}) \\[0.3em]
 H_{2t+1}(\mathbf{x}_{2t})
 \end{bmatrix}$
is $(2d)$-dimensional and wider.
(2) The output becomes narrower
after multiplication by
$\begin{bmatrix}
\mathbf{I} & \mathbf{I}
\end{bmatrix}$,
and the width is back to $d$.
(3) The block is indeed wider
except some trivial cases,
e.g., each residual branch does not contain nonlinear activations.

Figure~\ref{fig:wideinceptionlike} presents
an example to illustrate
that inception-like block is wider.
There are two layers in each branch.
We have that
the two residual branches is equivalent
to a single residual branch,
also containing two layers:
the first layer increases the width from $d$
($d=16$ in Figure~\ref{fig:wideinceptionlike})
to $2d$,
and the second layer
reduces the width back to $d$.
There is no such simple transformation for residual branches
with more than two layers,
but we have similar observations.

\vspace{.5em}
\noindent\textbf{Merge-and-run blocks are much wider.}
Consider Equation~\ref{eqn:mergeandrunblockforwidth},
we can see that
the widths of the input, the intermediate representation,
and the output are all $2d$\footnote{In essence,
the $2d$ space is not fully exploited
because the convolutional kernel is block-diagonal.}.
The block is wider than an inception-like block
because the outputs of two residual branches in the merge-and-run block are separated
and the outputs for the inception-like block are aggregated.
The two residual branches
are not independent
as the merge-and-run mapping
adds the input of one residual branch
to the output of the other residual branch.

\begin{figure}[t]
\centering
\footnotesize
{(a)~\includegraphics[scale=.45]{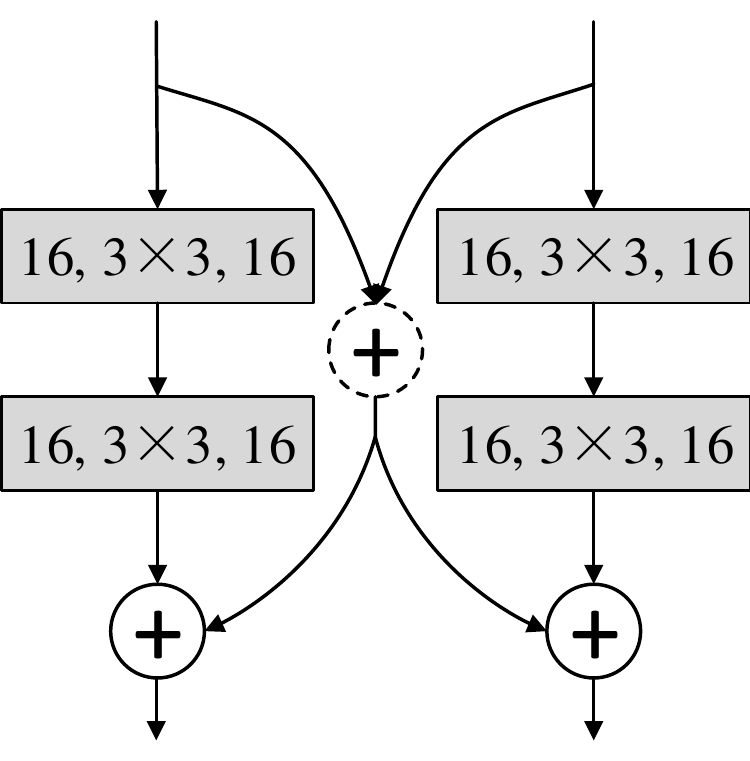}}~~~~~~
{(b)~\includegraphics[scale=.45]{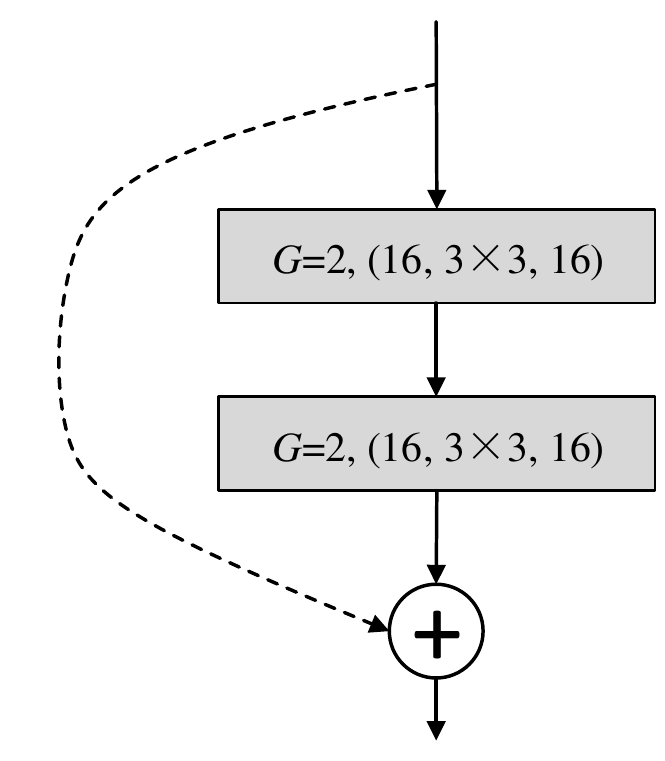}}
\caption{Transform the merge-and-run block shown in (a)
to a two-branch block shown in (b).
(b) The $2$ convolutions are group convolutions.
A group convolution contains two ($G=2$) convolutions of $(16, 3 \times 3, 16)$:
each receives a different $16$-channel input
and the two outputs are concatenated as the final output with $32$ channels.
The width is greater than $16$.
The skip connection (dot line) is a linear transformation,
where the transformation matrix of size $32\times 32$ is idempotent.}
\label{fig:widemergeandrun}\vspace{-.4cm}
\end{figure}

Figure~\ref{fig:widemergeandrun}
shows that
the merge-and-run block
is transformed to a two-branch block.
The dot line corresponds to
the merge-and-run mapping,
and now becomes an integrated linear transformation
receiving a single $(2d)$-dimensional vector as the input.
The residual branch consists of two group convolutions,
each with two partitions.
A group convolution is equivalent to a single convolution
with the larger convolution kernel,
being a block-diagonal matrix
with each block corresponding to the kernel of each partition
in the group convolution.

\section{Experiments}
We empirically show the superiority of
DILNets and DMRNets,
compared with ResNets.
We demonstrate the effectiveness
of our DMRNets on several benchmark datasets and compare
it with the state-of-the-arts.

\subsection{Datasets}
\noindent\textbf{CIFAR-$10$ and CIFAR-$100$.}
The two datasets are both subsets~\cite{KrizhevskyH09}
drawn from the $80$-million tiny image database~\cite{TorralbaFF08}.
CIFAR-$10$ consists of $60000$ $32\times 32$ colour images in $10$ classes,
with $6000$ images per class.
There are $50000$ training images and $10000$ test images.
CIFAR-$100$ is like CIFAR-$10$,
except that it has $100$ classes each containing $600$ images.
We follow a standard data augmentation scheme
widely used for this dataset~\cite{LeeXGZT14,ResNet,HuangLW16a}:
we first zero-pad the images
with $4$ pixels on each side,
and then randomly crop
to produce $32 \times 32$ images,
followed by horizontally mirroring
half of the images.
We preprocess the images
by normalizing the images
using the channel means and standard deviations.

\vspace{.5em}
\noindent\textbf{SVHN.}
The SVHN (street view house numbers) dataset
consists of digit images of size $32 \times 32$.
There are $73,257$ images as the training set,
$531,131$ images as a additional training set,
and $26,032$ images as the testing set.
Following the common practice~\cite{LinCY13,LeeXGZT14,StochasticDepth},
we select out $400$ samples per class from the training set and
$200$ samples per class from the additional set,
and use the remaining $598,388$ images for training.

\subsection{Setup}

\noindent\textbf{Networks.}
We follow ResNets to design our layers:
use three stages (conv$1\_x$, conv$2\_x$, conv$3\_x$) of merge-and-run blocks
with the number of filter channels
being $16,32,64$, respectively,
and use a Conv-BN-ReLU as a basic layer
with kernel size $3 \times 3$.
The image is fed into the first convolutional layer
(conv0)
with $16$ output channels,
which then go to the subsequent merge-and-run blocks.
In the experiments,
we implement our approach
by taking two parallel residual branches
as an example.
For convolutional layers with kernel size $3\times 3$,
each side of the inputs is zero-padded by one pixel.
At the end of the last merge-and-run block,
a global average pooling is performed and
then a soft-max classifier is attached.
All the $+$ operations in solid circles
in Figures~\ref{fig:networkblocks}
are between BN and ReLU.

\vspace{.5em}
\noindent\textbf{Training.}
We use the SGD algorithm with the Nesterov momentum
to train all the models for $400$ epochs on CIFAR-$10$/CIFAR-$100$
and $40$ epochs on SVHN,
both with a total mini-batch size $64$ on two GPUs.
The learning rate starts with $0.1$ and
is reduced by a factor $10$ at the $1/2$, $3/4$ and $7/8$ fractions
of the number of training epochs.
Similar to~\cite{ResNet,HuangLW16a},
the weight decay is $0.0001$,
the momentum is $0.9$, and the weights are initialized as in~\cite{MSRAinit}.
Our implementation is based
on MXNet~\cite{ChenLLLWWXXZZ15}.

\subsection{Empirical Study}

\begin{table*}[t]
\centering
\caption{Empirical comparison
of DILNets, DMRNets,
and ResNets.
The average classification error
from $5$ runs
and the standard deviation (mean $\pm$ std.)
are reported.
Refer to Table~\ref{tab:networtdescriptions}
for network structure descriptions.
}
\label{tab:comparewithresnetcifarsvhn}
\footnotesize{
\begin{tabular}{cc|ccc|ccc|ccc}
	\hline
	\multirow{2}[4]{*}{Params.} & \multirow{2}[4]{*}{$L$} & \multicolumn{3}{c|}{CIFAR-$10$}
	& \multicolumn{3}{c|}{CIFAR-$100$} & \multicolumn{3}{c}{SVHN}\\
	\cline{3-11}          &        & ResNets & DILNets & DMRNets  & ResNets & DILNets & DMRNets & ResNets & DILNets & DMRNets\\
	\hline
	$0.4$M  & $12$   & $6.62\pm0.24$ & $6.53\pm0.12$ & $\mathbf{6.48\pm0.04}$ & $29.69\pm0.15$ & $29.75\pm0.27$ & $\mathbf{29.62\pm0.08}$ & $\mathbf{1.90\pm0.08}$ & $2.13\pm0.09$ & $2.00\pm0.04$\\
	\hline
	$0.6$M  & $18$    & $5.93\pm0.17$ & $5.83\pm0.09$ & $\mathbf{5.79\pm0.13}$ & $27.90\pm0.26$ & $27.87\pm0.22$ & $\mathbf{27.80\pm0.26}$ & $1.97\pm0.09$ & $1.96\pm0.10$ & $\mathbf{1.87\pm0.09}$\\
	\hline
	$0.8$M  & $24$    & $5.60\pm0.14$ & $5.59\pm0.17$ & $\mathbf{5.47\pm0.14}$ & $27.03\pm0.66$ & $26.88\pm0.22$ & $\mathbf{26.76\pm0.16}$ & $1.93\pm0.17$ & $1.86\pm0.14$ & $\mathbf{1.86\pm0.05}$\\
	\hline
	$1.0$M  & $30$  & $5.50\pm0.09$ & $5.45\pm0.09$ & $\mathbf{5.10\pm0.08}$ & $26.44\pm0.69$ & $26.19\pm0.41$ & $\mathbf{25.87\pm0.04}$ & $1.89\pm0.03$ & $1.86\pm0.07$ & $\mathbf{1.81\pm0.04}$\\
	\hline
	$1.2$M  & $36$    & $5.35\pm0.14$ & $5.26\pm0.20$ & $\mathbf{5.18\pm0.20}$ & $26.00\pm0.48$ & $25.98\pm0.23$ & $\mathbf{25.41\pm0.19}$ & $1.90\pm0.04$ & $1.81\pm0.11$ & $\mathbf{1.77\pm0.11}$\\
	\hline
	$1.5$M  & $48$ &   $5.26\pm0.09$ & $5.05\pm0.20$ & $\mathbf{4.99\pm0.13}$ & $25.44\pm0.20$ & $24.76\pm0.33$ & $\mathbf{24.73\pm0.40}$ & $1.91\pm0.03$ & $1.84\pm0.06$ & $\mathbf{1.84\pm0.15}$\\
	\hline
	$1.7$M  & $54$ &  $5.24\pm0.23$ & $5.02\pm0.17$ & $\mathbf{4.96\pm0.06}$ & $24.56\pm0.15$ & $24.56\pm0.69$ & $\mathbf{24.41\pm0.09}$ & $2.00\pm0.12$ & $1.85\pm0.09$ & $\mathbf{1.68\pm0.02}$\\
	\hline
	$3.1$M  & $96$ & $5.47\pm0.46$ & $4.97\pm0.03$ & $\mathbf{4.84\pm0.11}$ & $24.41\pm0.10$ & $24.06\pm0.68$ & $\mathbf{23.98\pm0.15}$ & $1.85\pm0.03$ & $1.75\pm0.06$ & $\mathbf{1.70\pm0.03}$\\
	\hline
\end{tabular}%
}
\end{table*}

\noindent\textbf{Shorter paths.}
We study how the performance changes
as the average length of the paths changes,
based on two kinds of residual networks.
They are formed from the same plain network of depth $2L+2$,
whose structure is like the one forming ResNets
given in Table~\ref{tab:networtdescriptions}:
(i) Each residual branch is of length $\frac{2}{3}L$
and corresponds to one stage.
There are totally $3$ residual blocks.
(ii) Each residual branch is of length $\frac{1}{3}L$.
There are totally $6$ residual blocks (like Figure~\ref{fig:networkexamples} (a)).
The averages of the depths
of the paths are both $(L+2+1)$,
with counting two projection layers
in the shortcut connections.

\begin{figure}
  \centering
  \includegraphics[width=.9\linewidth]{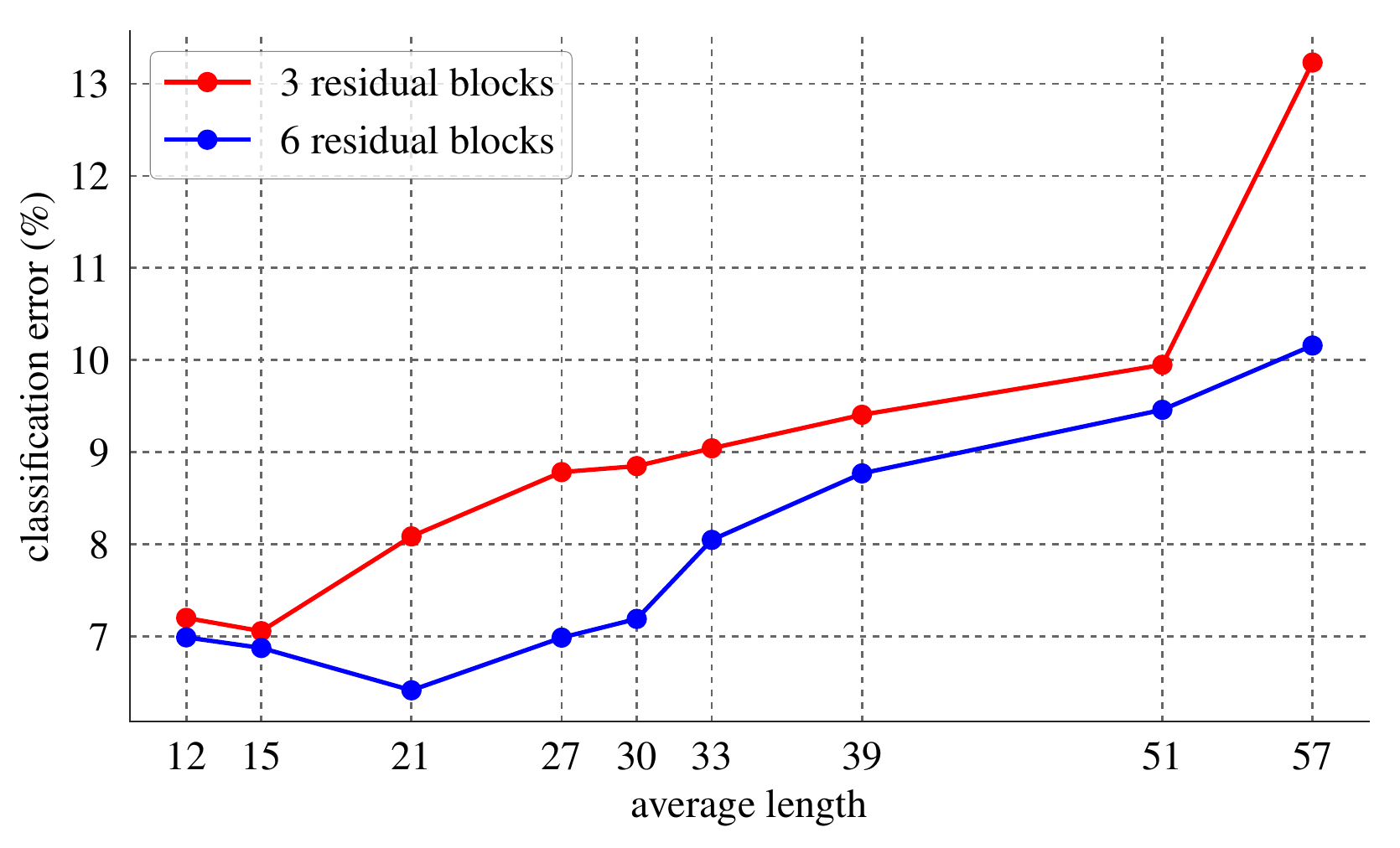}
  \caption{Illustrating
  how the testing errors
  of residual networks change
  as the average path length increases.
  The results are reported on CIFAR-$10$.}\label{fig:averagedepthvserrors}\vspace{-.4cm}
\end{figure}

We vary $L$ and record the classification errors
for each kind of residual network.
Figure~\ref{fig:averagedepthvserrors} shows the curves
in terms of the average depth of all the paths
vs. classification error
over the example dataset CIFAR-$10$.
We have the following observations.
When the network is not very deep and
the average length is small
($\leqslant 15$ for $3$ blocks,
$\leqslant 21 $ for $6$ blocks\footnote{There are more short paths for $6$ blocks,
which leads to lower testing error than $3$ blocks.}),
the testing error becomes smaller as
the average length increases,
and when the length is large,
the testing error becomes larger
as the length increases.
This indicates that
\emph{shorter paths}
result in the higher accuracy
for very deep networks.

\vspace{.5em}
\noindent\textbf{Comparison with ResNets.}
We compare DILNets and DMRNets,
and the baseline ResNets algorithm.
They are formed with the same number of layers,
and each block in a DILNet and a DMRNet
corresponds to two residual blocks
in a ResNet.
Table~\ref{tab:networtdescriptions}
depicts the network structures.

The comparison on CIFAR-$10$
is given in Table~\ref{tab:comparewithresnetcifarsvhn}.
One can see that compared with ResNets,
DILNets and DMRNets consistently
perform better, and DMRNets perform the best.
The superiority of DILNets over ResNets
stems from the less long paths and greater width.
The additional advantages of a DMRNet
are much greater width
than a DILNet.

The comparisons over CIFAR-$100$
and SVHN
shown in Table~\ref{tab:comparewithresnetcifarsvhn}
are consistent.
One exception is that
on CIFAR-$100$
the ResNet of depth $26$ ($L=12$)
performs better than DILNet
and
on SVHN the ResNet of depth $26$
performs better than
the DILNet and DMRNet.
The reason might be
that the paths in the DILNet and DMRNet
are not very long
and that too many short paths
lower down the performance
and that for networks of such a depth,
the benefit from increasing the width
is less than the benefit from increasing the depth.

\begin{figure*}[t]
	\centering
	\subfigure[CIFAR-$10$ ($L=30$)]{\label{fig:cifar10curve:MR}\includegraphics[width=0.33\linewidth, clip]{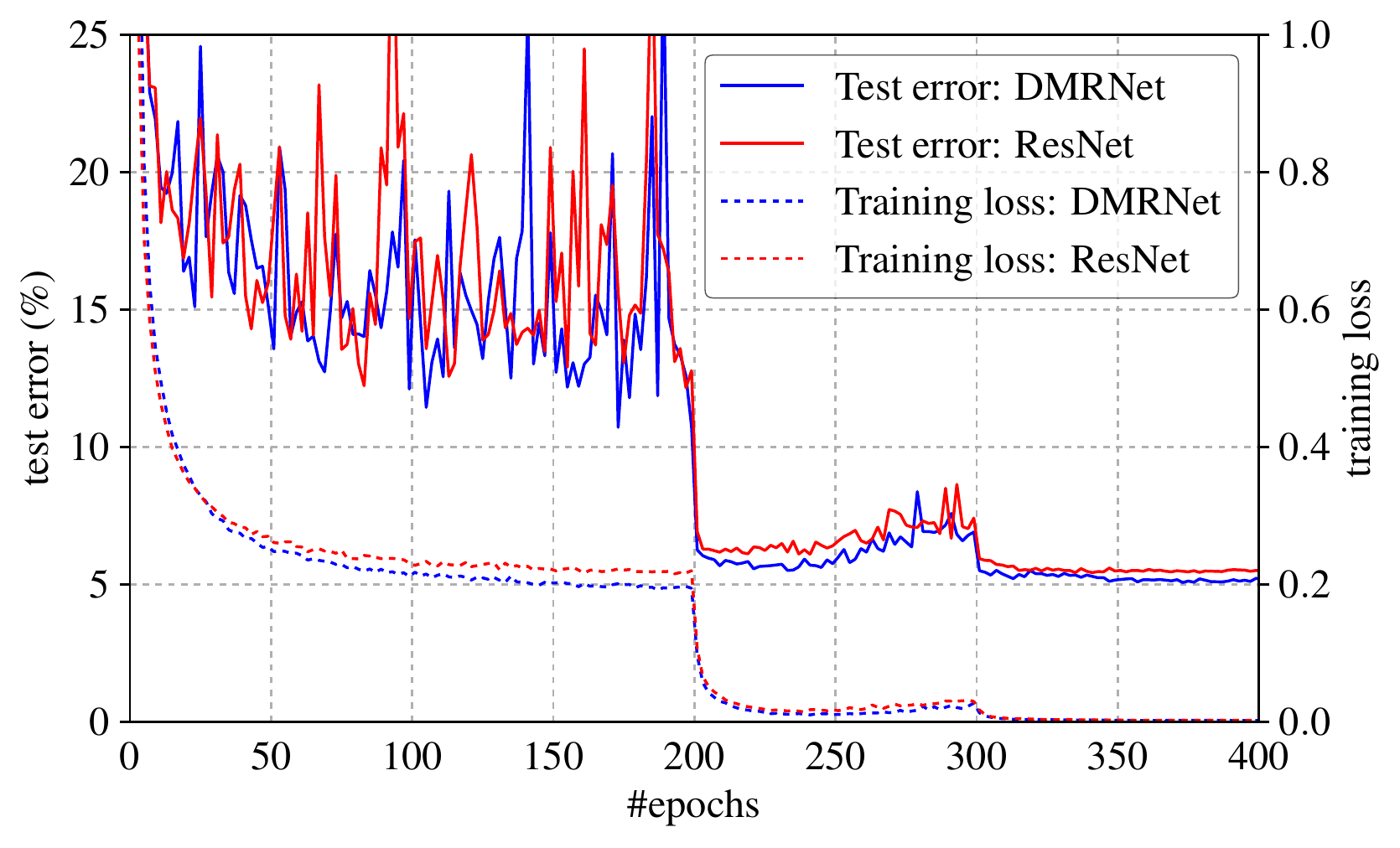}}~~\hfill~
	\subfigure[CIFAR-$100$ ($L=30$)]{\label{fig:cifar100curve:MR}\includegraphics[width=0.33\linewidth, clip]{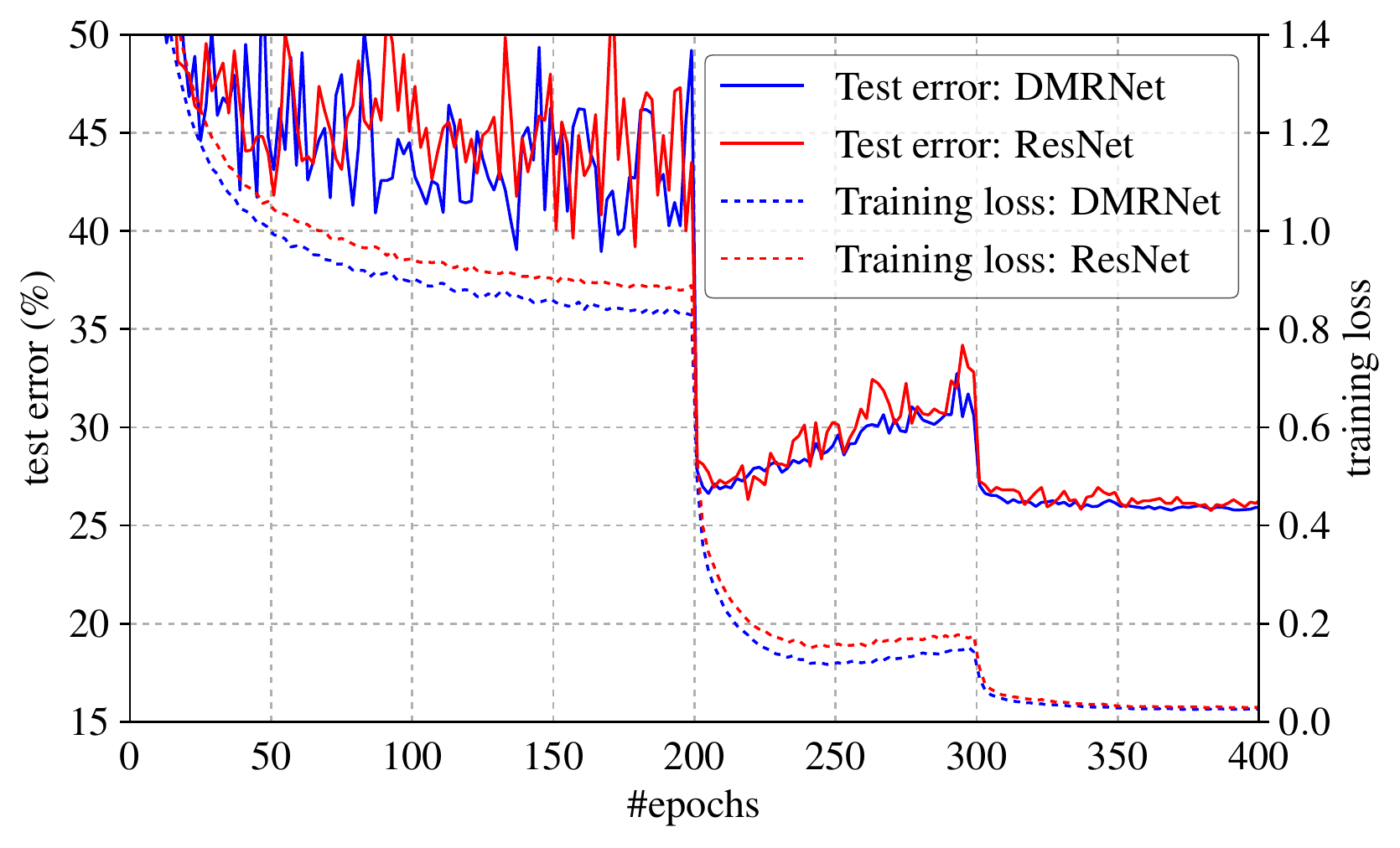}}~~\hfill~
	\subfigure[SVHN ($L=30$)]{\label{fig:svhncurve:MR}\includegraphics[width=0.33\linewidth, clip]{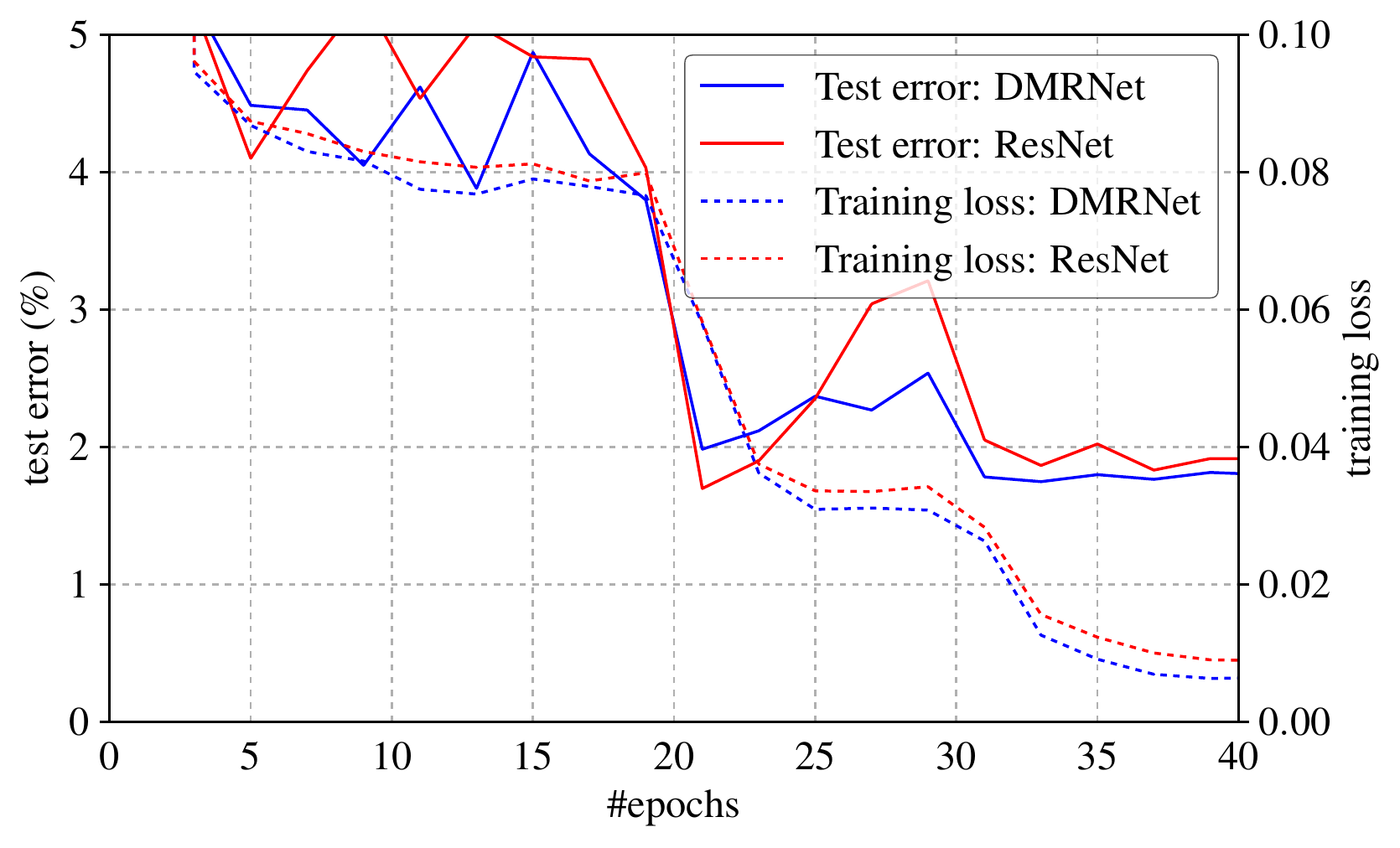}}
	\\
	\subfigure[CIFAR-$10$ ($L=54$)]{\label{fig:cifar10curve:MR}\includegraphics[width=0.33\linewidth, clip]{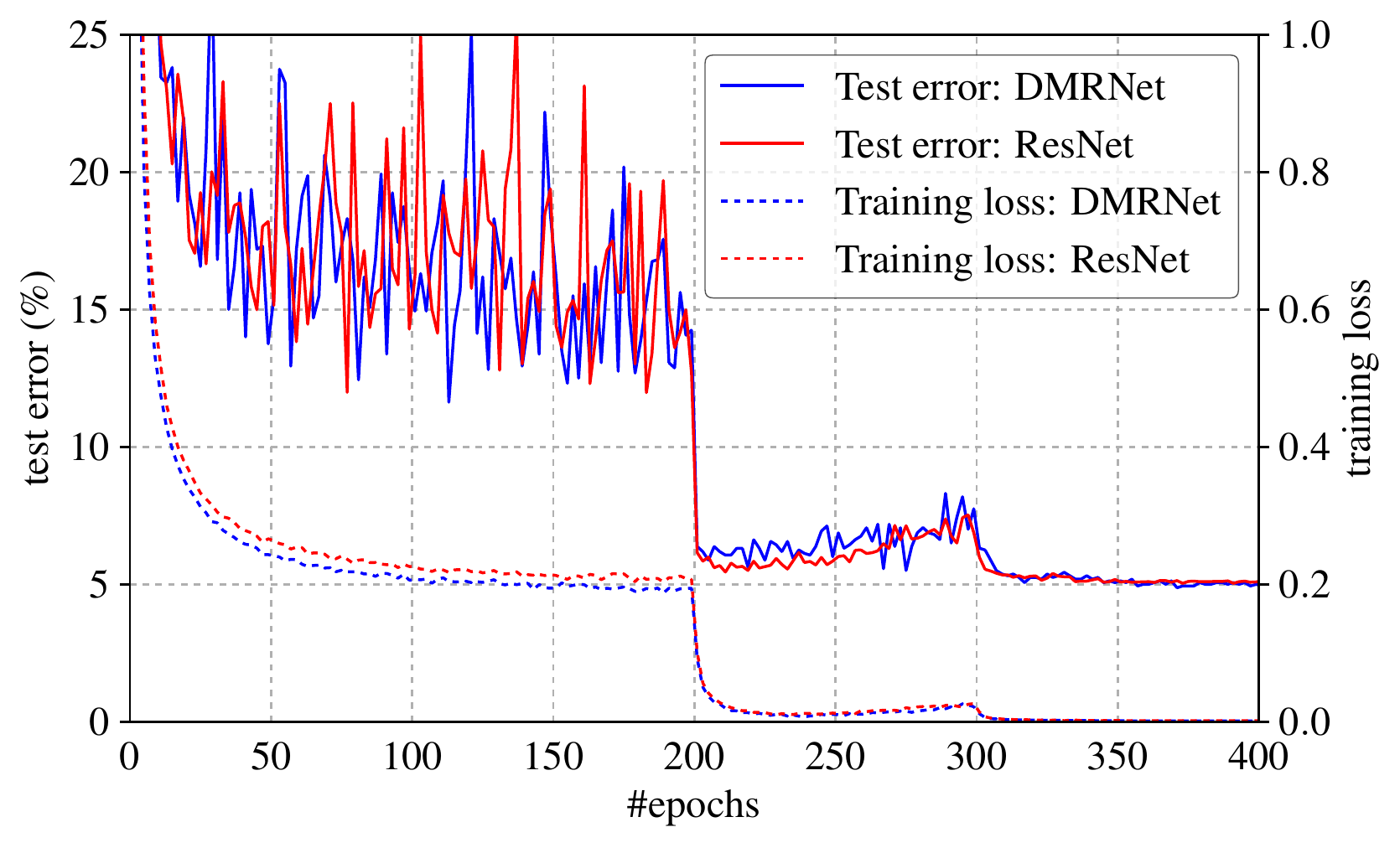}}~~\hfill~
	\subfigure[CIFAR-$100$ ($L=54$)]{\label{fig:cifar100curve:MR}\includegraphics[width=0.33\linewidth, clip]{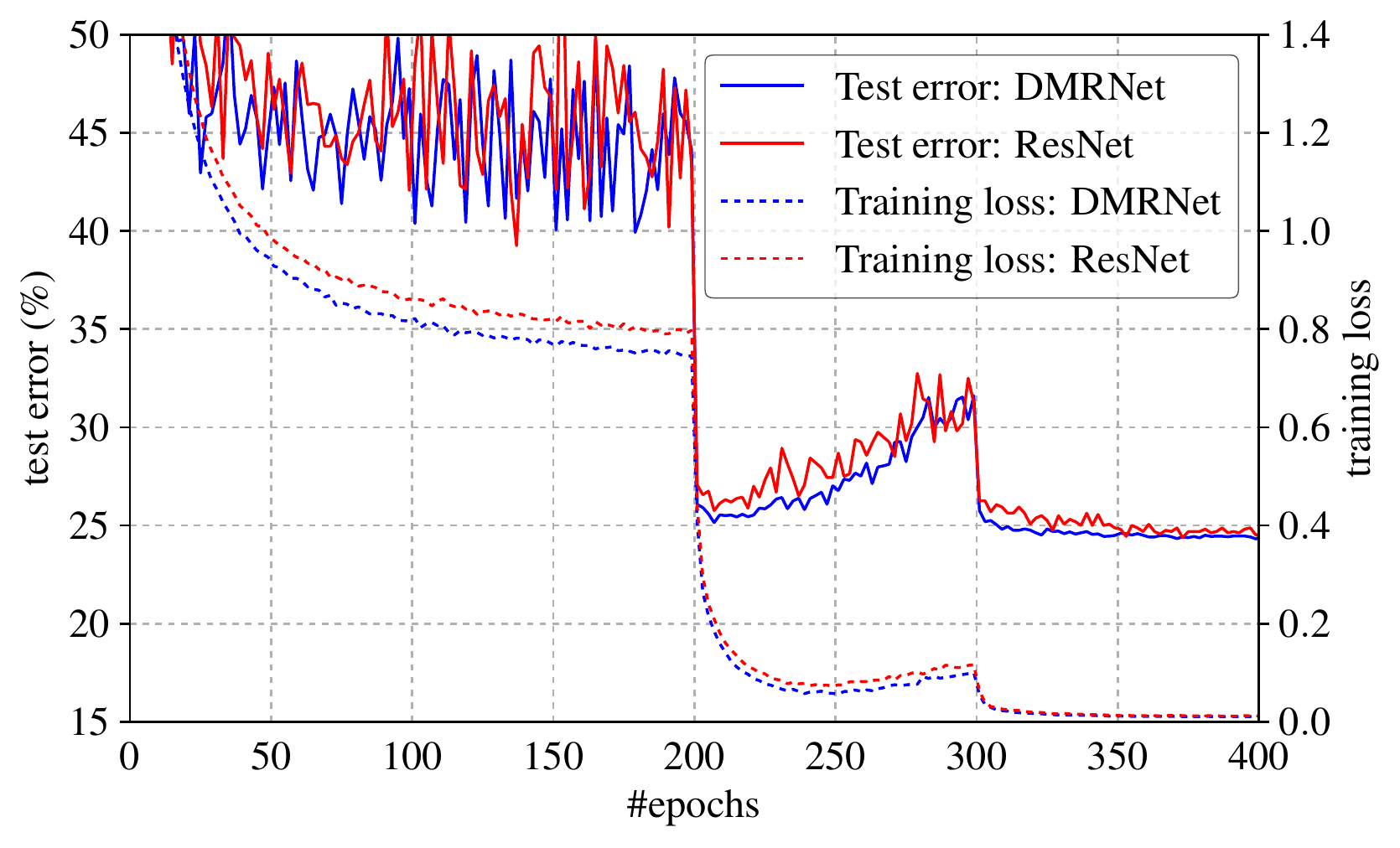}}~~\hfill~
	\subfigure[SVHN ($L=54$)]{\label{fig:svhncurve:MR}\includegraphics[width=0.33\linewidth, clip]{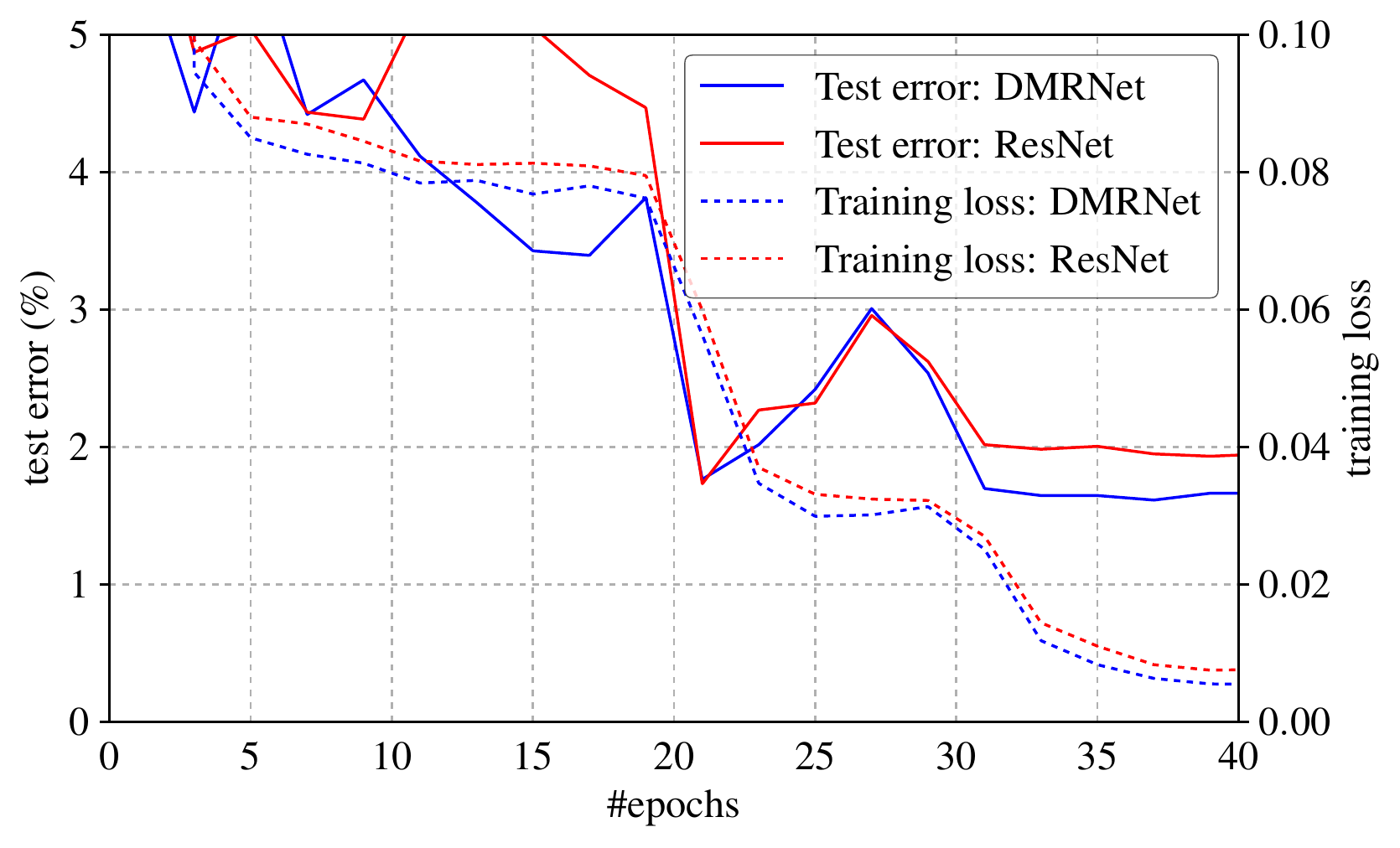}}
%
	\caption{Comparing the optimization
		of ResNets and the DMRNets
		with the same number of layers/parameters.
		The vertical axis corresponds to training losses
        and testing errors,
		and the horizontal axis corresponds to \#epochs.}
	\label{fig:convergecurve}\vspace{-.2cm}
\end{figure*}

\vspace{.5em}
\noindent\textbf{Convergence curves.}
Figure~\ref{fig:convergecurve} shows
the convergence curves
of ResNets and DMRNets
over CIFAR-$10$,
CIFAR-$100$, and SVHN.
We show training losses instead of training errors
because the training errors in CIFAR-$10$
almost reach zero at the convergence
and are not distinguishable.
One can see that the testing errors of DMRNets are smaller
than ResNets
and that the training losses
are also smaller
during the optimization process,
suggesting that our gains are not from regularization
but from richer representation.

\subsection{Comparison with State-of-the-Arts}

\begin{table*}[t]
\setlength{\tabcolsep}{12pt}
\centering
\caption{
	Classification error comparison
	with state-of-the-arts.
    The results of DenseNets
    are based on the networks without bottlenecks.
	The DMRNet-Wide is the wide version of a DMRNet,
    $4\times$ wider, i.e., the widths of the threes stages
    are $64$, $128$, and $256$, respectively.
    }
\label{tab:comparisonwithstateofthearts}%
\footnotesize
	\begin{tabular}{@{\extracolsep{\fill}}l|c | c | c | c | c}
		\hline
		& Depth & Params. & CIFAR-$10$ & CIFAR-$100$ & SVHN\\
		\hline
		Network in Network~\cite{LinCY13} &  -    &  -    & $8.81$  & -     & $2.35$\\
		All-CNN~\cite{allcnn2014} &  -    &  -    & $7.25$  & $33.71$ & - \\
		FitNet~\cite{fitnet2014} & -     & -     & $8.39$  & $35.04$ & $2.42$ \\
		Deeply-Supervised Nets~\cite{LeeXGZT14} &  -    &  -    & $7.97$  & $34.57$ & $1.92$ \\
		\hline
		Swapout~\cite{Swapout} &$ 20$    & $1.1$M & $6.85$  & $25.86$ & -\\
		&$ 32$    & $7.4$M & $4.76$  & $22.72$ & -\\
		\hline
		Highway~\cite{SimonyanZ14a} &  -    &  -    & $7.72$  & $32.39$ & -\\
		\hline
		DFN~\cite{WangWZZ16} &$ 50$    & $3.7$M & $6.40$  & $27.61$ & -\\
		&$ 50$    & $3.9$M & $6.24$  & $27.52$ & -\\
		\hline
		FractalNet~\cite{LarssonMS16a} &$ 21$    &  $38.6$M & $5.22$  & $23.30$ & $2.01$\\
		W/ dropout \& droppath &$ 21$    &  $38.6$M & $4.60$  & $23.73$ & $1.87$\\
		\hline
		ResNet~\cite{ResNet} & $110$   &  $1.7$M & $6.61$  & -     & -\\
		ResNet~\cite{StochasticDepth}  & $110$   &  $1.7$M & $6.41$  & $27.22$ & $2.01$\\
		ResNet (pre-activation)~\cite{ResNetV2} & $164$   &  $1.7$M & $5.46$  & $24.33$ & -\\
		& $1001$  &  $10.2$M & $4.62$  & $22.71$ & -\\
		\hline
		ResNet W/ stochastic depth~\cite{StochasticDepth} & $110$   &  $1.7$M & $5.23$  & $24.58$ & $1.75$\\
		& $1202$  &  $10.2$M & $4.91$  & -     & -\\
		\hline
		Wide ResNet~\cite{WideResNet} &$ 16$    &  $11.0$M & $4.81$  & $22.07$ & -\\
		&$ 28$    &  $36.5$M & $4.17$  & $20.50$ & - \\
		W/ dropout  &$ 16$    &  $2.7$M & -     & -     & $1.64$\\
		\hline
		RiR~\cite{TargAL16} & $18$   & $10.3$M& $5.01$  & $22.90$ & -\\
		\hline
		Multi-ResNet~\cite{MultiRes} & $200$   & $10.2$M& $4.35$  & $20.42$ & -\\
		& $398$   & $20.4$M& $3.92$  & -     & -\\
		\hline
		DenseNet~\cite{HuangLW16a}  
		& $100$   &  $27.2$M & $3.74$  & $19.25$ & $1.59$\\
		\hline
		DMRNet (ours) &$ 56$    & $1.7$M & ${4.96}$ & ${24.41}$  & $1.68$ \\
		DMRNet-Wide (ours) &$ 32$    & $14.9$M& $3.94$  & $\mathbf{19.25}$ & {\color{blue}$\mathbf{1.51}$} \\
		DMRNet-Wide (ours) &$ 50$    & $24.8$M& {\color{blue}$\mathbf{3.57}$} & {\color{blue}$\mathbf{19.00}$} & $\mathbf{1.55}$\\
		\hline
	\end{tabular}%
\vspace{-.2cm}
\end{table*}%

The comparison
is reported in Table~\ref{tab:comparisonwithstateofthearts}.
We report the results of DMRNets
since it is superior to DILNets.
Refer to Table~\ref{tab:networtdescriptions}
for network architecture descriptions.
We also report
the results from the wide DMRNets
(denoted by DMRNet-Wide),
$4\times$ wider, i.e.,
the widths of the threes stages
are $64$, $128$, and $256$, respectively.
We mark the results that outperform
existing state-of-the-arts in bold
and the best results in blue.

One can see that
the DMRNet-Wide of depth $50$
outperforms existing state-of-the-art results
and achieves the best results on CIFAR-$10$ and CIFAR-$100$.
Compared with the second best approach DenseNet{\footnote{
Both DMRNets and DenseNets do not use bottlenecks.
DenseNets with bottleneck layers would perform better.
We will combine bottleneck layers
into our approach
as our future work
to further improve the accuracy.}}
that includes more parameters ($27.2$M),
our network includes only $24.8$M parameters.
DMRNet-Wide (depth $=32$)
is very competitive:
outperform all existing state-of-the-art results
on SVHN.
It contains only $14.9$M parameters,
almost half of the parameters ($27.2$M)
of the competitive DenseNet.
These results show that
our networks are parameter-efficient.

Compared with the FractalNet with depth $21$,
DMRNets-Wide with depths $32, 50$
are much deeper
and contain fewer parameters ($14.9$M, $24.8$M vs. $38.6$M).
Our networks achieve superior performances over all the three datasets.
This also shows that because
\emph{merge-and-run mappings} improve information flow
for both forward and backward propagation,
our networks are less
difficult to train even though our networks are much deeper.


\subsection{ImageNet Classification}\label{sec:imagenetcurve}
We compare our DMRNet
against the ResNet on
the ImageNet $2012$ classification dataset~\cite{ImageNet},
which consists of $1000$ classes of images.
The models are trained on the $1.28$ million training images,
and evaluated on the $50,000$ validation images.

\vspace{0.1cm}
\noindent\textbf{Network architecture.}
We compare the results
of ResNet-$101$ with $101$ layers.
The ResNet-$101$~\cite{ResNet} ($44.5$M)
is equipped
with $4$ stages of residual blocks
with bottleneck layers,
and the numbers of blocks in the four stages
are $(3, 4, 23, 3)$, respectively.
We form our DMRNet
by replacing the residual blocks
with our merge-and-run
blocks and setting the numbers of blocks
in the four stages
to $(2,2,8,2)$,
resulting in our DMRNet with depth $44$ ($43.3$M).

\vspace{0.1cm}
\noindent\textbf{Optimization.}
We follow~\cite{ResNet}
and use SGD to train the two models
using the same hyperparameters
(weight decay $=0.0001$, and momentum $=0.9$)
with~\cite{ResNet}.
The mini-batch size is $256$,
and we use $8$ GPUs ($32$ samples per GPU).
We adopt the same data augmentation as in~\cite{ResNet}.
We train the models for $100$ epochs,
and start from a learning rate of $0.1$,
and then divide it by $10$ every $30$ epochs
which are the same as the learning rate changing in~\cite{ResNet}.
We evaluate on the single $224\times224$ center crop
from an image whose shorter side is $256$.

\vspace{0.1cm}
\noindent\textbf{Results.}
Table~\ref{tab:ImageNet101}
shows the results
of our approach and
our MXNet implementation of ResNet-$101$,
and the results of ResNet-$101$ from~\cite{ResNet}.
We can see that
our approach performs the best in terms of top-$5$ validation error:
our approach gets $1.7$ gain,
compared with the results of ResNet-$101$ from our implementation,
and $0.3$ gain
compared with the result from~\cite{ResNet}.

\begin{figure}[t]
	\centering
	\includegraphics[width=.9\linewidth]{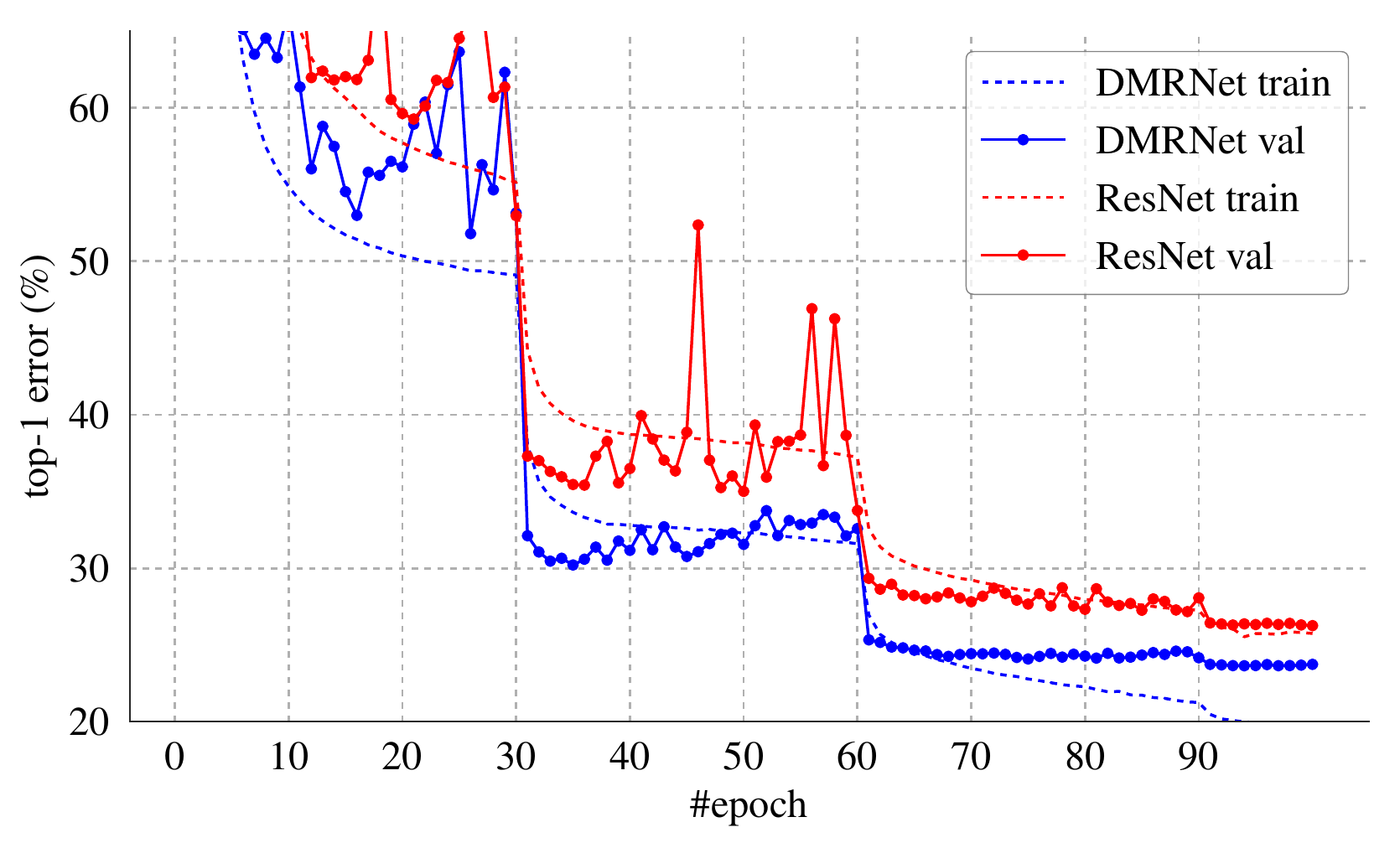}
	\caption{Training error and validation error
		curves of ResNet-$101$ ($44.5$M) and DFN-MR ($43.3$M)
		with the same optimization setting on ImageNet.
		We report the (top-$1$ error) results
        for training and
        single-crop validation.
		It can be observed that
		our approach performs better
		for both training errors and validation errors.}
	\label{fig:imagenet101}
	\vspace{-.2cm}
\end{figure}

The training and validation error curves of
ResNet-$101$ and our DMRNet
are given in Figure~\ref{fig:imagenet101}.
It can be observed that
our approach performs better
for both training errors and validation errors,
which also suggests that the gains are not from regularization
but from richer representation.
For example,
the top-$1$ validation error of our approach
is lower about $5\%$ than that of the ResNet
from the $30$th epoch to the $55$th epoch.

We notice that the results of our implemented ResNet on MXNet
and the results from~\cite{ResNet}
are different.
We want to point out that
the settings are the~\emph{same} with~\cite{ResNet}.
We think that the difference might be from
the MXNet platform,
or there might be some other untested issues
pointed by the authors of ResNets\footnote{\url{https://github.com/KaimingHe/deep-residual-networks}}.

\begin{table}[t]
	\centering
	\caption{The validation (single $224\times 224$ center crop)
        and training errors ($\%$)
		of ResNet-$101$ ($44.5$M) and our DMRNet ($43.3$M)
		on ImageNet.}
	\label{tab:ImageNet101}
\footnotesize
{
		\footnotesize
		\begin{tabular}{c|c c |c}
			\hline
			& ResNet-$101$~\cite{ResNet} & ResNet-$101$ & DMRNet \\
			\hline
			\#parameters &  \multicolumn{2}{c|}{$44.5$M}  & $43.3$M \\
			\hline
			Top-$1$ validation error  & $23.60$ & $26.41$  & $23.66$   \\
			Top-$5$ validation error  & $7.10$ & $8.50$   & $\mathbf{6.81}$   \\
			\hline
			Top-$1$ training  error & $17.00$ & $25.75$  & $19.72$ \\
			Top-$5$ training error  & - & $8.12$   & $6.59$   \\
			\hline
		\end{tabular}
	}
\vspace{-.3cm}
\end{table}

\section{Discussions}
\label{sec:discussion}

\noindent\textbf{Merge-and-run mappings for $K$ branches.}
The merge-and-run mapping studied in this paper
is about two residual branches.
It can be easily extended to
more ($K$) branches,
and accordingly merge-and-run mappings
become a linear transformation
where the corresponding transformation matrix
is of $K \times K$ blocks,
with each block being $\frac{1}{K}\mathbf{I}$.

\vspace{.5em}
\noindent\textbf{Idempotent mappings.}
A merge-and-run mapping
is a linear
idempotent mapping,
which is a linear transformation
with the transformation matrix $\mathbf{M} $ being idempotent,
$\mathbf{M}^n = \mathbf{M}$.
Other idempotent mappings can also be applied
to improve information flow.
For examples, the identity matrix $\mathbf{I}$
is also idempotent
and can be an alternative
to the merge-and-run mappings.
Compared with identity mappings,
an additional advantage is
that merge-and-run mappings introduce
interactions between residual branches.

We conducted experiments
using a simple idempotent mapping, $\mathbf{I}$,
for which there is no interaction
between the two residual branches
and accordingly the resulting network
consists of two ResNets that are separate
except only sharing
the first convolution layer
and the last FC layer.
We also compare the performances
of the two schemes without sharing those two layers.
The overall superior results of our approach,
from Table~\ref{tab:comparemergeandrunandidentity},
show that the interactions introduced by merge-and-run mappings
are helpful.

\begin{table}
\centering
\caption{Comparison between merge-and-run mappings
and identity mappings. Sharing = share the first conv. and the last FC.}
\footnotesize
\label{tab:comparemergeandrunandidentity}
\begin{tabular}{c|c|c|c|c|c}
\hline
&& \multicolumn{2}{c|}{CIFAR-$10$} & \multicolumn{2}{c}{CIFAR-$100$} \\
\hline
&L & Identity & Merge-and-run & Identity & Merge-and-run \\
\hline
\multirow{2}{*}{w/ sharing} &$48$ & $5.21$ & $\mathbf{4.99}$ & $25.31$ & $\mathbf{24.73}$ \\
& $96$ & $5.10$    & $\mathbf{4.84}$ & $24.16$     & $\mathbf{23.98}$ \\
\hline
\multirow{2}{*}{w/o sharing} & $48$ & $4.67$ & $\mathbf{4.41}$ & $23.96$ & $\mathbf{23.75}$ \\
& $96$ & $4.51$    & $\mathbf{4.37}$ & $\mathbf{22.23}$     & ${22.62}$ \\
\hline
\end{tabular}
\vspace{-.3cm}
\end{table}

Our merge-and-run mapping is complementary
to other design patterns,
such as dense connection in DenseNet~\cite{HuangLW16a},
bottleneck design, and so on.
As our future work,
we will study the integration with other design patterns.

%
%

\vspace{.5em}
\noindent\textbf{Deeper or wider.}
Numerous studies have been conducted
on going deeper,
learning very deep networks,
even of depth $1000+$.
Our work can be regarded as
a way to going wider and less deep,
which is also discussed in~\cite{WuSH16e,WideResNet}.
The manner of increasing the width in our approach is different
from Inception~\cite{GoogLeNet},
where the outputs of the branches are concatenated for~\emph{width increase}
and then a convolution/pooling layer for each branch in the subsequent Inception block
is conducted but for {width decrease}.
Our merge-and-run mapping suggests a novel and cheap way
of increasing the width.

\section{Conclusions}
In this paper,
we propose deep merge-and-run neural networks,
which improves residual networks
by assembling residual branches in parallel
with merge-and-run mappings
for further reducing the training difficulty.
The superior performance stems
from several benefits:
Information flow is improved,
the paths are shorter,
and the width is increased.

{\small
\bibliographystyle{ieee}
\bibliography{ensemble}
}

\end{document}